\documentclass[12pt]{article}

\usepackage{scicite}
\usepackage{times}
\usepackage{pbox}
\usepackage{latexsym}
\usepackage[pdftex]{graphicx}
\usepackage{hyperref}
\usepackage{booktabs}
\usepackage{subfigure}
\usepackage{amsmath}
\usepackage{amsbsy}
\usepackage{cases}
\usepackage{bm}
\usepackage{algorithm}
\usepackage{algorithmic}
\usepackage{multirow}

\newcounter{lastnote}

\title{SlangSD: Building and Using a Sentiment Dictionary of Slang Words for Short-Text Sentiment Classification}

\author
{Liang Wu, Fred Morstatter, Huan Liu\\
\\
\normalsize{Arizona State University}\\
\normalsize{Tempe, Arizona, 85287-8809}\\
\normalsize{ \{wuliang, fred.morstatter, huan.liu\}@asu.edu }
}

\date{}

\begin{document}


\baselineskip24pt


\maketitle


\begin{abstract}
Sentiment in social media is increasingly considered as an important resource for customer segmentation, market understanding, and tackling other socio-economic issues. However, sentiment in social media is difficult to measure since user-generated content is usually short and informal. Although many traditional sentiment analysis methods have been proposed, identifying slang sentiment words remains untackled. One of the reasons is that slang sentiment words are not available in existing dictionaries or sentiment lexicons. To this end, we propose to build the first sentiment dictionary of slang words to aid sentiment analysis of social media content. It is laborious and time-consuming to collect and label the sentiment polarity of a comprehensive list of slang words. We present an approach to leverage web resources to construct an extensive Slang Sentiment word Dictionary (SlangSD) that is easy to maintain and extend. SlangSD is publicly available\footnote{http://www.slangsd.com/} for research purposes. We empirically show the advantages of using SlangSD, the newly-built slang sentiment word dictionary for sentiment classification, and provide examples demonstrating its ease of use with an existing sentiment system.
\end{abstract}

\section{Introduction}\label{secintro}
The massive amount of information from general users on microblogging platforms provides valuable insights. One of the ways these insights are obtained is by analyzing the sentiment of content generated by users. However, a challenging issue for measuring sentiment in user-generated content is the detection of opinionated expressions. For example, after Apple launched their iPhone 6S in September of 2015, comments were posted such as ``Apple you knocked it out of the park!'', and ``battery life's shit hot''. The slang words/phrases ``out of the park'' and ``shit hot'', which are used to express the feeling of ``best'' and ``excellent'', literally seem to be negative instead. Therefore, a fundamental task of analyzing user-generated sentiment expressions is to identify the polarity of these slang words.

In order to efficiently identify sentiment words, several existing sentiment lexicons have been built. The construction of a sentiment lexicon consists of two steps, collecting subjective words and assigning sentiment polarities. Since their vocabulary mostly depends on lexical resources for analyzing formal content, existing lexicons are not capable of measuring slang sentiment words. For example, the commonly used lexicons including SentiWordNet~\cite{baccianella2010sentiwordnet}, Micro-WNOp~\cite{cerini2007micro}, and the lexicon used in \cite{hu2004mining}\footnote{https://www.cs.uic.edu/~liub/FBS/sentiment-analysis.html\#lexicon}, all adopted the vocabulary of WordNet~\cite{miller1995wordnet}, where slang words and phrases are not present.

In this work, we introduce the first sentiment dictionary of slang words. Building an extensive sentiment dictionary is challenging since (1) Most slang words are not present in existing sentiment lexicons, so collecting a comprehensive vocabulary is difficult; and (2) Classifying sentiment polarity is very time-consuming and labor-intensive. In order to tackle the first challenge, we propose to utilize web resources. Websites were developed to collect the meanings of slang words such as Urban Dictionary\footnote{http://www.urbandictionary.com/}. As a result of active volunteer participation, they have a complete and up-to-date slang word list. We propose to exploit the slang word vocabulary from Urban Dictionary (UD), which is the leading website among slang word dictionary sites according to Alexa\footnote{http://www.alexa.com/topsites/category/Top/Reference/Dictionaries}.

Although UD nicely provides a list of slang words, the sentiment polarity remains to be discovered. In order to tackle the challenge of obtaining sentiment polarity, we propose to leverage existing sentiment lexicons, social media corpora, and the synonyms of slang words from UD, to automatically estimate the sentiment polarity. Details of sentiment polarity classification are discussed in Section \ref{sec:est}. We will also present preliminary results on integrating SlangSD into mining sentiment from user-generated content. Our main contributions of the work can be summarized as follows:
\begin{itemize}
\item{Build the first sentiment dictionary for slang words. The first version of this dictionary contains 96,462 slang words and phrases with sentiment scores. We make it publicly available,}
\item{Introduce a principled way of automatically labeling the sentiment polarity of slang words at a large scale, and}
\item{Demonstrate the utility of SlangSD by using it in an existing sentiment classification system with real-world data sets of short and informal text.}
\end{itemize}

\section{Related Work}

Since a good sentiment lexicon is crucial to identifying sentiment expressions, such as SMS messages and tweets~\cite{nielsen2011new}, various methods have been proposed over the past few years to generate lexicons. For example, sentiment strengths can be propagated with synonymous relationships between words. The relationship can be synsets in WordNet~\cite{miller1995wordnet,hu2004mining}, syntactic patterns in corpora~\cite{qiu2011opinion}, and contextual information in sentences~\cite{kanayama2006fully}. Moreover, building domain-specific lexicons~\cite{hai2014identifying} and leveraging crowdsourced annotations~\cite{mohammad2013crowdsourcing} have been studied in previous work. Other methods directly model sentence sentiments~\cite{pang2002thumbs,socher2013recursive}.

Our work is also related to generating lexical databases. The most commonly used English lexical dictionary is WordNet~\cite{miller1995wordnet}, where words, meanings, relationships are well compiled in the structured database. Similar words are clustered together as different synsets. As mentioned before, WordNet can be used for generating a sentiment dictionary. A sentiment word database called SentiWordNet~\cite{baccianella2010sentiwordnet} was developed by classifying the sentiment strength of WordNet synsets. Commonly used sentiment dictionaries also include Harvard Inquirer\footnote{http://www.wjh.harvard.edu/~inquirer/spreadsheet\_guide.htm}, Micro-WNOp~\cite{cerini2007micro}, MPQA~\cite{deng2015mpqa}, LIWC~\cite{pennebaker2001linguistic}, VADER~\cite{hutto14parsimonious}, the lexicon used in \cite{hu2004mining}\footnote{https://www.cs.uic.edu/~liub/FBS/sentiment-analysis.html}, etc. Other methods focus on structures and compositions of sentences~\cite{pang2002thumbs,socher2013recursive}.

Existing efforts on identifying and measuring sentiment slang words mainly focus on specific corpus, by leveraging the context of slang words. For example, ``LOL'' is found to appear frequently around ``funny'', so ``LOL'' probably means causing amusement~\cite{pak2010twitter}. Tang et al. obtain a seed set of slang words from UD to help generate representation for words in Twitter~\cite{tang2014building}. These methods are limited by the completeness of the corpus, which usually generate domain-specific lexicons and require retraining on more data to cater to a new task. To the best of our knowledge, we are the first to leverage Urban Dictionary to build an extensive slang word sentiment dictionary for analyzing sentiment in short and informal user-generated content.%

Urban Dictionary has been used to analyze slang words. The vocabulary can help identify slang words in online reviews~\cite{gruhl2010multimodal}. Chen et al. further identify sentiment expressions by using UD as a resource of slang word vocabulary~\cite{chen2012extracting}. Our work is different from theirs, since we further leverage different resources to estimate the sentiment strength for slang words from UD, instead of merely using the vocabulary.

Table~\ref{comparison} illustrates the differences between our SlangSD and other related lexical resources. Formal words are those which appear in dictionaries such as WordNet. Slang words (phrases) are those which are not present in dictionaries, while are widely used for expressing sentiment. As shown in the table, existing sentiment lexicons mainly focus on formal words, which do not contain an extensive list of slang words. UrbanDictionary has an extensive list of slang words, while the sentiment polarity is unavailable. SlangSD is the first sentiment lexicon, which simultaneously covers a comprehensive list of slang words and the sentiment polarity.

\begin{table}
	\centering
	\caption{Comparison of related lexical resources in terms of vocabulary and sentiment polarity. }
	\begin{tabular}{c c c c }
		\toprule
	& Formal Words & Slang Words & Sentiment \\
		\toprule
		\pbox[c][26pt][b]{3.9cm}{ Existing Lexicons\\  \cite{baccianella2010sentiwordnet,deng2015mpqa,cerini2007micro,pennebaker2001linguistic,hutto14parsimonious,hu2004mining}} & Extensive & Incomplete & Available\\
		\toprule
		WordNet~\cite{miller1995wordnet} &Extensive & Unavailable & Unavailable\\
		\toprule
		UrbanDictionary &Incomplete & Extensive & Unavailable\\
		\toprule
		SlangSD &Incomplete & Extensive & Available\\
		\bottomrule
	\end{tabular}
	\label{comparison}
\end{table}

\section{SlangSD: A Sentiment Dictionary of Slang Words}

In this section, we introduce how we collect slang words from UD, and how we estimate the sentiment strength of slang words.
\subsection{Collecting Slang Words}
\textbf{Urban Dictionary} is a crowdsourced online dictionary of slang words founded in 1999, where the words, their meaning, and all other items are uploaded by online volunteers. A word may have multiple meanings, so the order of being displayed relies on the user votes. In summary, we crawled 5 items of a slang word including \textbf{meaning}, \textbf{example sentences}, \textbf{related words}, \textbf{upvote}, \textbf{downvote}. Note that all listed slang words have at least one meaning and one example sentence, while some of them are without related words and have few votes. Since UD was famous for its satirical characteristic, many meanings are uploaded sarcastically. For example ``LoL'', of which the funny meaning (abbreviation of Laurence) is the most popular\footnote{http://www.urbandictionary.com/define.php?term=Lol}. Therefore, we propose to only exploit the related words for sentiment classification. Next we will introduce details about how we leverage available resources to solve the problem.

\subsection{Estimating Sentiment Strength}\label{sec:est}

In estimating the sentiment strength of slang words, we scale the strength from -2 to 2, where -2 is strongly negative, -1 is negative, 0 is neutral, 1 is positive, and 2 is strongly positive.

\textbf{Sentiment lexicon:} First, since a small portion of slang words from UD have also been defined by existing sentiment lexicons, such as SentiWordNet~\cite{baccianella2010sentiwordnet}, LIWC~\cite{pennebaker2001linguistic}, MPQA~\cite{deng2015mpqa}, and the sentiment lexicon compiled by earlier work~\cite{hu2004mining}, we leveraged these existing definitions to label the corresponding words. For example, we labeled the words obtained from UD that also appear in existing sentiment lexicons, such as ``hilarious'' and ``gross'', with their sentiment strength in existing lexicons. For words which have different sentiment strengths in different lexicons, we adopt the average. By checking the four aforementioned dictionaries, we found sentiment strengths for \textbf{761} words collected from UD. However, this accounts for less than 1\% of the dictionary. Next, we introduce how we leverage the usage of slang words in social media content to estimate the sentiment strength.

\textbf{Twitter as a labeling mechanism:}
As discussed in Section~\ref{secintro}, word co-occurrences can be used to estimate the sentiment strength. The underlying assumption is that words that co-occur frequently are likely to have similar meanings. Therefore, we propose to leverage Twitter, since slang words are widely used in tweets. Twitter provides a Search API\footnote{https://dev.twitter.com/rest/public/search/}, by which tweets containing the exact word(s) and/or phrase(s) can be quickly retrieved. In particular, we use a slang word as a query, and randomly retrieve up to 150 tweets with the API. Then the sentiment strength can be estimated by the average strength of sentiment words that are closest to the query in each tweet~\cite{turney2002unsupervised}. Note that a tweet without a listed sentiment word is assumed to be neutral (with the sentiment strength of 0). There are \textbf{22,710} slang words that can be labeled by the corpus of Twitter, which account for an additional 23\%. Next, we introduce how we further expand the vocabulary of SlangSD by exploiting UD features.
\begin{figure}[!t]
\centering
\includegraphics[width=3.1 in]{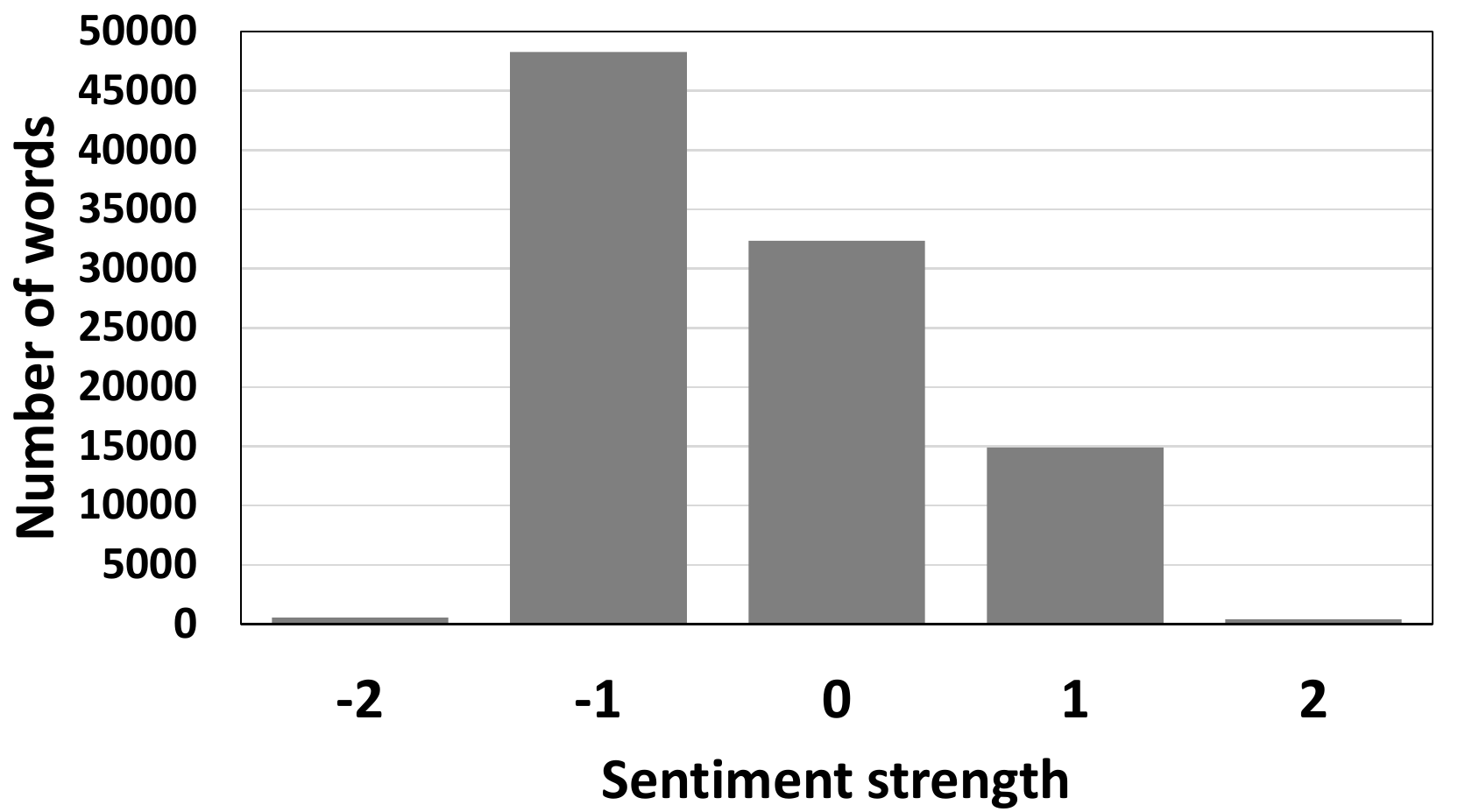}
\caption{Distribution of sentiment strength. The largest sentiment group is weakly negative (-1), where 50\% of all slang words on UD fall in it.}
\label{sentiment_distribution}
\end{figure}

\textbf{Sentiment propagation:}
A list of related words may be available for slang words in UD. Such synonymous relationships have been studied to infer sentiment polarity~\cite{hu2004mining,baccianella2010sentiwordnet}, since words with similar meanings are likely to share the same sentiment polarity. In particular, we use the slang words with sentiment strength as the seed set, and then annotate the connected unlabeled words. The average is adopted if an unlabeled word has related words with conflicting sentiment strengths. We iterate until the sentiment strength is no longer updated. \textbf{72,991} words are labeled this way, which accounts for an additional 76\%. We finally came up with a vocabulary of \textbf{96,462} slang words with sentiment strength. The distribution of sentiment polarity can be found in Figure~\ref{sentiment_distribution}. The largest sentiment strength group is -1.

\subsection{Extending SlangSD}
New slang words are created continually. Therefore, it is essential to keep extending the vocabulary of SlangSD. Extending SlangSD consists of two steps: obtaining new slang words and labeling their sentiment polarity; both steps can be automated. One way to get newly-generated slang words is to query UD based on the creation date for new words. For example, slang words created on July 14$^{th}$, 2016 can be retrieved via querying UD by visiting   ``http://www.urbandictionary.com/yesterday.php\\?date=2016-07-14''. Slang words can be thus regularly or periodically retrieved by altering  the date part, ``2016-07-14'', in the url.

As discussed in Section~\ref{sec:est}, three methods are available for automatically labeling the sentiment polarity of a slang word. For slang words present in existing sentiment lexicons, the sentiment polarity can be directly obtained for these slang words. The second method is to labeling new slang words by identifying their co-occurring known sentiment words in the same posts. To retrieve a post we can leverage the corpus of Twitter by using its Search API\footnote{https://dev.twitter.com/rest/public/search/} for retrieving content containing the slang word.  The third method is to leverage relationships between words as synonyms of a slang word should share similar sentiment. Since UD attaches a list of related words for most slang words, we can obtain synonyms of slang words and thus, label the sentiment of a slang word by the known sentiment of its synonyms.

\section{Utility of SlangSD}\label{sec:method}
In this section, we investigate whether SlangSD is useful for  identifying sentiment from user-generated content in practice. Next, we conduct experiments to assess its utility on real datasets.

\begin{table}
	\centering
	\caption{Statistics of datasets. }
	\begin{tabular}{c c c c c}
		\toprule
	& Instances & Slang & Positive & Negative\\
		\toprule
		SMS &2,000 & 645 & 286& 253 \\
		\toprule
		Twitter &1,598,962 & 572,641 & 220,928& 225,364 \\
		\bottomrule
	\end{tabular}
	\label{datastat}
\end{table}

\subsection{Twitter and SMS Data}
Informal text is pervasively present in online and daily communication. In order to evaluate SlangSD, we use tweets from Twitter and messages from Short Message Service (SMS). The Twitter data is collected from Twitter's Streaming API\footnote{https://dev.twitter.com/streaming/reference/post/statuses/filter}. This API returns a subset of tweets randomly. As a common practice in labeling tweets at a large scale~\cite{ramage2010characterizing,agarwal2011sentiment}, we use a predetermined set of emoticons\footnote{https://en.wikipedia.org/wiki/List\_of\_emoticons} to tag the sentiment strength of tweets. The SMS dataset is sampled from SemEval 2013 dataset\footnote{http://alt.qcri.org/semeval2014/task9/}, which is publicly available. The dataset contains sentences extracted from short message services which are labeled with sentiment polarity.

Statistics of the datasets are shown in Table~\ref{datastat}. SMS is a small dataset, which contains 2,000 messages. 645 messages contain words listed in SlangSD, of which 286 are with a positive sentiment polarity, 253 negative, and the rest neutral. The Twitter dataset is much bigger.
\subsection{Experimental Settings}

\begin{table}
	\centering
	\caption{Accuracy of different models.}
	\begin{tabular}{ l l l}
		\toprule
		Method& Twitter & SMS\\
		\toprule
		DeeplyMoving & 56.17\% &66.28\%\\
		SentiStrength& 65.08\% &73.15\%\\
		SentiStrength$_{\textbf{SSD}}$& \textbf{84.84\%} &\textbf{86.55\%}\\
		\bottomrule
	\end{tabular}
	\label{resulttotal}
\end{table}
Sentiment analysis methods can be generally categorized into sentence-structure-based methods and lexicon-based methods. Sentence-structure-based methods focus on composition of words~\cite{socher2013recursive} and lexicon-based methods focus on finding key subjective expressions~\cite{thelwall2010sentiment}, such as sentiment words/phrases and negations, to determine the polarity. Two corresponding algorithms below are chosen for comparison purposes.

\textbf{DeeplyMoving} is a state-of-the-art deep learning method to identify sentiment strength in sentences~\cite{socher2013recursive}. DeeplyMoving focuses more on sentence structures and has been widely used for mining opinions.

\textbf{SentiStrength}~\cite{thelwall2010sentiment} is a state-of-the-art lexicon-based method for identifying sentiment in user-generated content.

SlangSD would be most useful for sentiment lexicon-based methods as an additional dictionary of slang sentiment words. Thus, we augment SentiStrength with SlangSD to construct a third method next.

\textbf{SentiStrength$_{\textbf{SSD}}$} is SentiStrength incorporated with sentiment words from SlangSD. We add words in SlangSD into the original sentiment word lexicon.

Through experiments, we evaluate the effectiveness of SlangSD by comparing  SentiStrength with SentiStrength$_{\textbf{SSD}}$ and we also examine their performance relative to that of DeeplyMoving.

\subsection{Sentiment Analysis Results}

\begin{table}
	\centering
	\caption{Results on Tweets containing slang.}
	\begin{tabular}{l l l l l}
		\toprule
		&&Precision&Recall&F-score\\
		\toprule
\multirow{2}{*}{Positive} & SentiStrength & 65.30\% & 78.13\% &	71.14\%	\\
&SentiStrength$_{\textbf{SSD}}$ & 65.69\% & 79.86\% &	72.09\%	\\
		\toprule

\multirow{2}{*}{Negative}&SentiStrength & 66.25\% & 76.62\% &	71.06\%	\\
&SentiStrength$_{\textbf{SSD}}$ & 89.91\% & 92.50\% &	91.19\%	\\
		\bottomrule
	\end{tabular}
	\label{resulttwitter}
\end{table}

\begin{table}
	\centering
	\caption{Results on SMS messages containing slang.}
	\begin{tabular}{l l l l l}
		\toprule
		&&Precision&Recall&F-score\\
		\toprule
\multirow{2}{*}{Positive} & SentiStrength & 71.08\% & 20.63\% &	31.98\%	\\
&SentiStrength$_{\textbf{SSD}}$ & 77.42\% & 75.51\% &	76.46\%	\\
		\toprule

\multirow{2}{*}{Negative}&SentiStrength & 75.58\% & 25.69\% &	38.35\%	\\
&SentiStrength$_{\textbf{SSD}}$ & 79.93\% & 88.14\% &	83.83\%		\\
		\bottomrule
	\end{tabular}
	\label{resultsms}
\end{table}
The performance of the three different models is illustrated in Table~\ref{resulttotal}. It can be seen from the results that DeeplyMoving achieves the lowest accuracy among all three. The SentiStrength model outperforms DeeplyMoving. It focuses more on identifying subjective expressions in short and informal text. Through incorporating SlangSD, SentiStrength$_{\textbf{SSD}}$ achieves the best result among the three methods.

In order to further explain how SlangSD facilitates the problem of sentiment analysis on content containing slang words, we report the results on tweets and messages containing slang words, which are shown in Table~\ref{resulttwitter} and Table~\ref{resultsms}, respectively. The one-vs-all setting is adopted for calculating the Precision, Recall and F-score. It can be seen that, SentiStrength$_{\textbf{SSD}}$ performs much better in terms of F-score on negative tweets. Since UD is well known for satire, the results show that SlangSD may well enrich the lexicon on negative sentiment. On the SMS dataset, SentiStrength$_{\textbf{SSD}}$ outperforms SentiStrength by over 50\% recall on both tasks. Since SentiStrength was mostly applied in the area of social media, the corresponding lexicon may not cover enough sentiment words in messages. SlangSD complements the lexicon to be more extensive. In conclusion, the results indicate that SlangSD offers a list of slang sentiment words, which help identify subjective expressions from short and informal text in different areas, and are more crucial for interpreting negative sentiment. Since UD is famous for satire, the words may naturally be more useful for interpreting negative than positive.

\begin{figure}[!t]
\centering
\includegraphics[width=3.3in]{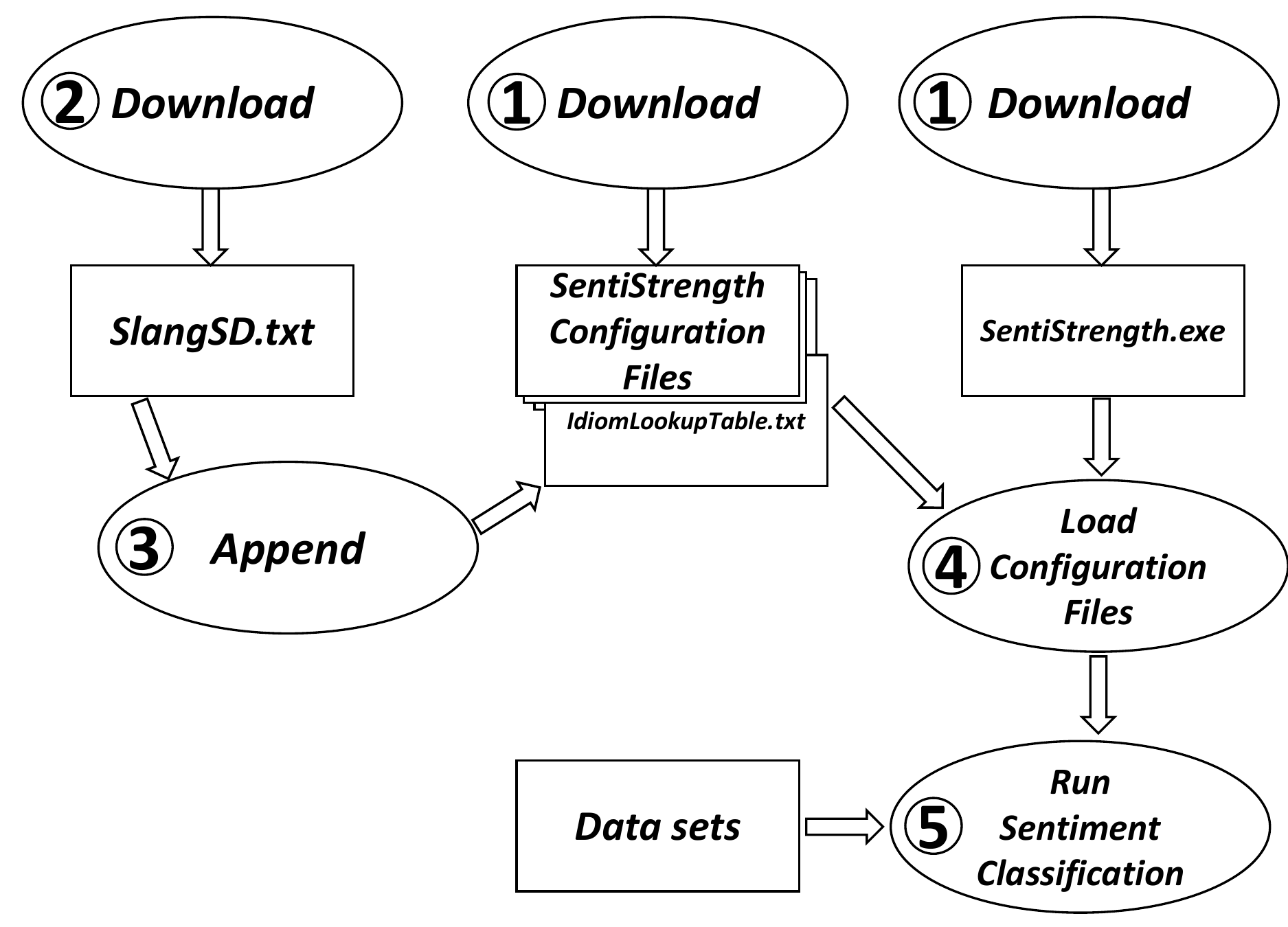}
\caption{Process of incorporating SlangSD with SentiStrength.}
\label{process}
\end{figure}
\section{An Example Illustrating how to use SlangSD}
In this section, we introduce how to incorporate SlangSD with SentiStrength. We illustrate the process briefly in Figure \ref{process}. The first step is to download the configuration files and the executable file of the software\footnote{http://sentistrength.wlv.ac.uk/download.html}. The configuration files include several dictionaries for SentiStrength. In order to exploit the additional slang words, second we download the dictionary file ``SlangSD.txt''\footnote{http://slangsd.com/data/SlangSD.zip} and append it to ``IdiomLookupTable.txt''. We suggest to use ``IdiomLookupTable.txt'' since many phrases are present in SlangSD. These phrases share the similar format of idioms (a combination of words). The fourth step is to load the configuration files. Then SentiStrength can be used to classify sentiment of the data sets.

If the 3$^{rd}$ and 4$^{th}$ steps are skipped, the results are based on original lexicons without SlangSD, which have been used as baselines in Section~\ref{sec:method}. Details about configuring and running SentiStrength are available on the SentiStrength website\footnote{http://sentistrength.wlv.ac.uk/\#About}.
\section{Conclusion}\label{secconc}

Sentiment analysis in short and informal text is a fundamental problem for various domains. Although methods are proposed to solve this problem, a key challenge of identifying sentiment in informal/short text is the lack of lexical resources for understanding the sentiment strength of slang words. To this end, we propose a web-search-based, learning approach to build the first slang sentiment word dictionary, named SlangSD, by leveraging the available online resources. It is shown that SlangSD can effectively improve the state-of-the-art informal text sentiment analysis tool, and it can be easily incorporated as an additional sentiment lexicon. The future work includes continuing adding new slang words to broaden the coverage of SlangSD, and maintaining the vocabulary of SlangSD via updates and assessment. The publicly available SlangSD will enable collective efforts to use SlangSD in various sentiment analysis tasks.

\section*{Acknowledgments}

We would like to thank DMML members for their feedback and help in this work. The work is funded, in part, by ONR N00014-16-1-2257 and the Department of Defense under the MINERVA initiative through the ONR N00014131083. 

\bibliography{slangsd}

\bibliographystyle{Science}


\end{document}